# Semantic Segmentation of Radar Detections using Convolutions on Point Clouds

**M. Braun**[1, 2, a], **A. Cennamo**[1, 2, b], **M. Schoeler**[1, c], **K. Kollek**[2, d] **and A. Kummert**[2, e]

[1] Aptiv Services Deutschland GmbH, Wuppertal, 42119, Germany  
[2] University of Wuppertal (BUW), Wuppertal, 42119, Germany

Email: [a]marco.braun@aptiv.com; [b]alessandro.cennamo@aptiv.com; [c]markus.schoeler@aptiv.com; [d]kevin.kollek@uni-wuppertal.de; [e]kummert@uni-wuppertal.de

**Abstract.** For autonomous driving, radar sensors provide superior reliability regardless of weather conditions as well as a significantly high detection range. State-of-the-art algorithms for environment perception based on radar scans build up on deep neural network architectures that can be costly in terms of memory and computation. By processing radar scans as point clouds, however, an increase in efficiency can be achieved in this respect. While Convolutional Neural Networks show superior performance on pattern recognition of regular data formats like images, the concept of convolutions is not yet fully established in the domain of radar detections represented as point clouds. The main challenge in convolving point clouds lies in their irregular and unordered data format and the associated permutation variance. Therefore, we apply a deep-learning based method introduced by PointCNN that weights and permutes grouped radar detections allowing the resulting permutation invariant cluster to be convolved. In addition, we further adapt this algorithm to radar-specific properties through distance-dependent clustering and pre-processing of input point clouds. Finally, we show that our network outperforms state-of-the-art approaches that are based on PointNet++ on the task of semantic segmentation of radar point clouds.

## 1. Introduction

Modern driver assistance systems, up to fully autonomous driving, require a reliable perception of the vehicle's environment. To ensure this, vehicles have been equipped with sensors such as camera, lidar and radar. The data from those sensors are then processed to derive scene information. Machine Learning (ML)-based algorithms, in particular, are successfully used to process camera images by applying Convolutional Neural Networks (CNNs) [1] [2]. Unlike the regular data format of camera images, however, lidar and radar sensors produce point clouds (PC). A PC is defined as an unordered collection of $N \in \mathbb{N}$ individual points with coordinates $p_i \in \mathbb{R}^D$, $i = 1, …, N$, in $D \in \mathbb{N}$ dimensions, each of which is assigned a feature vector $f_i \in \mathbb{R}^F$, $i = 1, …, N$, $F \in \mathbb{N}$. CNNs, however, cannot be efficiently applied to PCs due to their irregular and unordered data format. Various approaches [3] [4] [5] aim to solve this problem.

Radar sensors play a major role in autonomous driving vehicles due to their high reliability in challenging weather conditions as well as a wide detection range. PCs returned by radar sensors, however, are sparse and have a lower spatial accuracy compared to lidar data, potentially impeding the task of extracting features from shape information. On the other hand, radar measurements provide intrinsic properties like the relative, radial speed of objects, also known as Doppler velocity $v_r$, and the radar cross section ($\sigma$) as a measure for the reflectivity.

In this work, we focus on perceiving the environment of a vehicle based on radar reflections. By integrating radar reflections over time, approaches like occupancy grid maps extract static information around to the ego car [6]. These approaches, however, fail to account for dynamic objects. On the other hand, state-of-the-art tracking algorithms used to perceive motion of objects are computationally expensive. By processing radar detections individually, this issue can be avoided. Therefore, we present a novel approach to perform semantic segmentation on radar PCs, i.e. classifying each prediction individually. We use a ML-based algorithm to derive information from a combination of intrinsic properties of radar detections like doppler velocities along with spatially local correlations.

Recent ML state-of-the-art approaches process PCs using PointNet [3] and PointNet++ [4]. We instead decide to use PointCNN [5] as our key feature extractor. PointCNN essentially differs from PointNet and PointNet++ in the way it maintains permutation invariance while processing PCs: Approaches that build up on PointNet use a maxpooling operation on individually extracted features to maintain permutation invariance during processing clusters of points. In contrast, by building up on PointCNN, we use a neural network to weight and permute features of clustered points to obtain permutation invariance. This operation then makes the application of classical CNNs on the feature maps of grouped radar detections possible. Furthermore, we use a pre-processing network (PP) presented in [7] to derive high level patterns by combining the coordinates *(x, y)* with $v_r$ and $\sigma$ of each detection individually and thus optimize the initial representation of radar-related features. We observed, that radar-detections in our dataset are distributed very heterogeneously on the x-y plane. This property potentially has a negative impact on the efficient extraction of information from spatially local correlations of detections. Therefore, we propose to apply multi-scale-grouping (MSG) [4] to cluster detections instead of the k-nearest neighbour algorithms applied in PointCNN. Finally, we show that our approach outperforms the state-of-the-art algorithm PointNet++ as a basic building block for semantic segmentation on radar PCs.

## 2. Related work

A broad variety of algorithms like SegNet [8], U-Net [9], DeepLabv3+ [10] and HRNet [11] show superior performance in semantic segmentation of images by building up on CNNs [1]. In order to apply those approaches on PCs, points need to be transferred into a homogenous data format similar to pixels in images. Once the PC is represented in a grid map format, 2D/ 3D CNNs can be applied for feature extraction. Building up on that, shape recognition and semantic segmentation is carried out in [12] [13] by processing PCs into grid cells at different angles and then merging the extracted features in 3D. Although these approaches represent a method of adapting CNNs to point clouds, a grid resolution must be defined that exponentially scales the computational operations that are needed for data processing. Therefore, by transferring PCs to a grid map, a compromise must be made between computational effort and loss of structural information by applying a coarse grid.

Alternatively, algorithms [3] [4] [5] [14] [15] [16] were devised to consume PCs without the need of processing them to a 2D/ 3D grid. A central contribution on directly processing PCs is PointNet [3]. By recursively applying shared multi-layer perceptrons (MLP) [17] on each point in a PC individually, permutation invariance is conserved while the algorithm extracts high-level descriptive patterns. This approach is extended in PointNet++ [4] by clustering detections into overlaying sub-regions depending on their relative distance and applying PointNet [3] on each cluster individually. Thus, the network is capable of extracting spatially local correlations to recognize fine-grained patterns. PointNet++ [4] follows an encoder - decoder structure expanded by skip-links between hierarchical related layers: Encoding of cluster-related characteristics into representative points is recurrently performed in set-abstraction (SA) layers. In this part of the network, the amount of points in the processed PCs decreases while each remaining point contains a more and more expressive signature of the original PC. The highly descriptive PC resulting from SA layers is then decoded in feature-propagation (FP) layers. These layers propagate high level features back to the point locations of the initial PC. Skip connections are used to fuse extracted features of FP and SA layers and thus enrich the expressiveness of the resulting PC.

Schumann et al. [18] applied PointNet++ to perform semantic segmentation on radar PCs. They convey detections into a two-dimensional environment *p = (x, y)* with individual features *f = (x, y, $v_r$, σ)*. As a result, individual probabilities are predicted for each detection of the processed radar PC for dynamic classes *car*, *truck*, *pedestrian*, *pedestrian group*, *bike* or *static*. Moreover, Feng et al. applied a slightly modified version of PointNet++ on Radar PCs to discretize between vehicles and guardrails [19]. During SA, they calculate statistics such as density of a cluster and append these characteristics to the corresponding local detections. Further approaches from Wöhler et al. [20] used a sampling algorithm like DBSCAN [21] to group together points of the same object and then applied a long short term memory (LSTM) [22] to perform semantic segmentation of the resulting grid. The authors prove the superior performance of the Deep Learning (DL) based LSTM algorithm compared to random forest algorithm [23].

## 3. Method

State-of-the-art approaches that perceive the environment of a vehicle by classifying radar PCs [7] [18] build up on PointNet++ [4] as a central feature extractor. PointNet++, however, aims to shift the idea of CNNs to the domain of PCs by individually extracting patterns from radar detections within a PC and then forwarding the most expressive features of neighbouring points to the next layer. In this way, permutation invariance is conserved at the expense of inefficient extraction of features. Due to these compromises, previous PointNet++ based approaches show structural weaknesses in pattern recognition from spatial-local correlations compared to conventional CNNs. By utilizing PointCNN [5], we circumvent these issues by pre-processing points within a cluster of radar detections, aiming to obtain a consistent ordering. A classical convolution can then be carried out on the invariant order of points achieved in this way. The authors of PointCNN demonstrate the superior performance of their approach compared to PointNet and PointNet++ for processing sparse PCs. Radar PCs, which are examined in this work, are characterized by this sparsity. By using PointCNN as the basic feature extractor, we therefore present a novel and potentially superior approach for the semantic segmentation of radar PCs.

The success of modern ML-based approaches for classification and semantic segmentation lies in the extraction of spatially local correlations in the data [2]. Therefore, we transform our obtained radar detections to a two-dimensional coordinate system *(x, y)* with radar-related features $v_r$ and σ. Then, we define a proportion of radar detections as representative points (RP). These RP then represent regions of interest we want to extract features from. RP are obtained by applying a sampling algorithm like farthest points sampling (FPS) on the PC. Afterwards, we associate radar detections to each RP based on their Euclidian distance by using a grouping mechanism like k-nearest neighbors. RPs, along with their assigned detections, then represent clusters from which we want to derive patterns. As described in previous sections of this work, the direct convolution on those clusters poses challenges due to the irregular and unordered data format of PCs. The authors of PointCNN address this issue by introducing a so-called X-transformation on the detections of each cluster we want to extract descriptive patterns from and call the resulting layer X-Convolution.

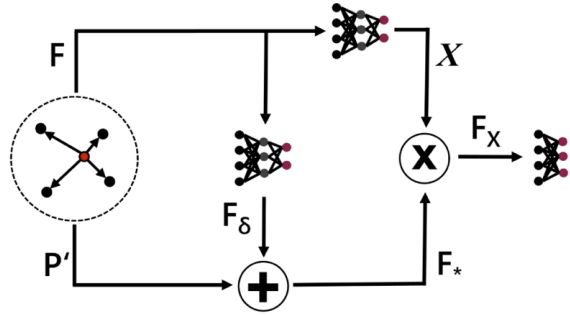

**Figure 1.** X-Convolution: For each cluster of radar detections, we learn a X-Transformation matrix to weight and permute features of the gathered points to obtain permutation invariance. A classical convolution can then be performed on those pre-processed features.

According to [5], Figure 1 shows the calculations within a X-Convolution that we apply to pre-process each cluster of radar detections to obtain permutation invariance. First, we move the coordinates $p$ of radar detections within a cluster to a coordinate system around their corresponding RP so that the RP is located at $p_{RP} = (0, 0)$, by performing

$$P' = P - p \qquad (1)$$

with P representing the coordinates of RP. We then lift those local coordinates $P'$ into high level features $F_\delta$ by feeding them individually into a shared $MLP_\delta$.

$$F_\delta = MLP_\delta(P') \qquad (2)$$

The resulting, coordinate-dependent features $F_\delta$ of each radar detection are then concatenated with the features F of each point in the cluster to get $F_*$.

$$F_* = [F_\delta, F] \qquad (3)$$

Finally, we want to encode relations between radar detections within the cluster. At this point, we use the property of PointCNN to achieve permutation invariance for unordered points within a cluster. We now use a MLP to extract a K x K matrix by processing P', where K is the number of points within a cluster.

$$X = MLP(P') \qquad (4)$$

The K x K matrix is called X-Transformation matrix. This resulting X-Transformation matrix depends on the order of detections. According to PointCNN this is desired, since we then apply the X-Transformation matrix to simultaneously weight and permute features $F_*$. The order of detections in $F_*$ is now related to the ordering of local coordinates we use to calculate the X-Transformation matrix. Therefore, weighting and permuting $F_*$ by X by performing a matrix product potentially leads to a permutation invariant feature mapping $F_X$.

$$F_X = X \cdot F_* \qquad (5)$$

$F_X$ can then be used as an input to a classical convolution. Descriptive patterns that result from X-Convolutions are then aggregated on the RP of each cluster, respectively. With each layer of X-Convolutions, the size of the resulting point cloud is therefore reduced to the number of RPs. Also, each of those resulting points contains encoded patterns from grouped radar detections of the previous layer.

Our network follows the encoder-decoder structure. In the encoder part, we recursively apply X-Convolutions and therefore receive decreasingly dense PCs, while each remaining point aggregates extensive patterns from a subsequently enlarging receptive field. Accordingly, as a result of the final encoding layer, we receive a PC that contains a minority of points, each of which aggregates highly descriptive contextual information. These high-level features then need to be propagated back to the high resolution of the initial PC. We achieve this by re-applying X-Convolutions recursively. This time, however, the resulting point cloud contains more points after each layer, which reflects the size of the respective PCs in the encoding part of the network. In addition, during decoding, we use skip links from encoded layers of the same depth to concatenate features. Finally, we use those resulting feature vectors from our decoded, high resolution PC to predict classification scores for each radar detection. Therefore, we apply a shared MLP to process high level feature vectors of each detection individually.

PointCNN was originally developed to process dense laser PCs. Radar data, however, is characterized by its low density as well as its high spatial inaccuracy while each radar reflection contains valuable intrinsic properties. We therefore propose some modifications to better adapt PointCNN to the processing of radar data: First, we deploy a shared MLP on the input feature vector containing $x$, $y$, $v_r$ and $\sigma$ as proposed by [7]. In this way, the network can convert radar-specific properties into an extended representation that facilitates the semantic segmentation task. In additions, radar PCs are distributed very heterogeneously throughout the scene. By applying k-nearest neighbour algorithm to group together neighbouring detections, this circumstance leads to clusters of widely spread detections. Features that we obtain by performing X-Convolutions on those clusters do not necessary incorporate spatially local correlations anymore. As a second modification, we therefore deploy multi-scale grouping (MSG) introduced by PointNet++ [4] to improve clustering of our approach. In MSG, we specify a number of points that should be clustered as well as the maximum distance of those detections to their corresponding RP. If there are not enough detections within the given radius, we randomly double detections until we reach the desired cluster size. Thus, we can be sure that grouped detections are always in close proximity to one another.

## 4. Experiments

*4.1 Data*

In our work we perform semantic segmentation on three classes: *moving vehicle*, *moving pedestrian* and *static* objects. Detections were assigned to a class by the help of annotated bounding boxes. Due to the spatial inaccuracy of radar detections, we increased the scope of the bounding boxes and assigned the class of a bounding box to each detection within its boundaries. Objects were defined as moving if their absolute velocity was above the value of 2.5m/s for vehicles and 0.5 m/s for pedestrians. Statistics on the distribution of classes in our data set can be obtained from Table 1. To test how well our model generalizes on unseen data, we split our data set into 80% for training and 20% for testing.

**Table 1.** Data Split: The relative (below: absolute) occurrence of the individual classes in the test and training data set.

| Class | Vehicle | Pedestrian | Static |
|---|---|---|---|
| **Train** | 6.12% | 2.26% | 91.62% |
|  | 1 955 937 | 721 865 | 29 282 877 |
| **Test** | 5.63% | 1.40% | 92.98% |
|  | 574 321 | 142 647 | 9 493 667 |

The radar PCs that we process for training and testing our approach initially consist of coordinates $p_0 = [x, y]$ and features $f_0 = [v_r, \sigma]$. One sample on average consists of 184 detections with a variance of 79 detections. Due to the sparsity of our data, we concatenated and ego-motion compensated four consecutive PCs. We aligned the position of all four PCs to the coordinate system of the last frame.

We set the size of our assembled PCs to 1200 detections. In case the number of detections was below this threshold, we doubled random detections, in case the amount was above, we randomly subsampled detections.

*4.2 Training*
We train our network described in section 3 for semantic segmentation by backpropagation [24]. In order to indicate the confidence score for classes *moving vehicle* and *moving pedestrian* we apply a classification head with two output channels. To ensure a range between 0 and 1 for each confidence score, we apply a sigmoid function to the logits of the network outputs. We then distinguish between the two classes and the *static background* as follows: If, for a given prediction for a specific detection no confidence score is above a threshold of 0.5, this detection is predicted as *background*. If at least one of the confidence values is greater than 0.5, the detection is classified as the class with the highest confidence value.

Due to the huge imbalance between classes in our data as depicted in Table 1, we used Focal Loss [25] with $\alpha_{vehicle} = 0.8$ and $\alpha_{pedestrian} = 0.95$ together with $\gamma = 2$. The training was carried out by Adam Optimizer [26] with a learning rate of $10^{-4}$. We trained each network for 20 epochs and used those with the highest $F_1$-Score [27] for comparison. No data augmentation was applied.

*4.3 Results*
In this section we present and discuss the ability of our approach to individually classify radar detections. A visual analysis of our network performance for the class *moving vehicles* can be obtained from Figure 2. It shows a single frame of radar reflections from our test set in an urban environment. While most detections are correctly classified as *moving vehicles* or *static background*, some false positive moving vehicle predictions can occasionally be observed. This type of misclassification can be traced back to sensor-related noises caused by time synchronization or mirroring effects.

The scene, however, illustrates the capability of our approach to distinguish between detections from the same object versus background noise. Furthermore, the visualization of a radar PC in Figure 2 shows the heterogenous distribution of detections within the scene which, as mentioned in Section 3, motivates us to replace k-nearest neighbour algorithm with MSG to form clusters within the PC.

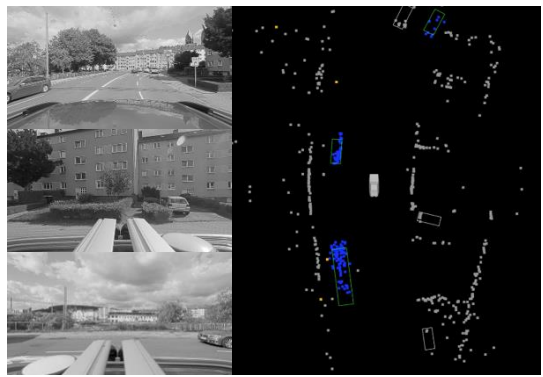

**Figure 2.** Analysis of our approach in an urban environment consisting of blue (true positive - TP), white (true negative - TN) and yellow (false negative – FN) moving vehicle predictions. Green bounding boxes indicate a dynamic and white a stationary object.

Before we quantitatively compared our approach with previous algorithms, we carried out experiments to find an optimal spatial representation of the input PCs. As stated above, our initial PC consists of coordinates *x* and *y* together with features $v_r$ and $\sigma$. Experiments, however, showed that applying $v_r$ as an additional coordinate, e.g. $p_0 = [x, y, v_r]$ improved grouping of detections which resulted in a better overall performance. Accordingly, we applied the doppler signature of any radar detection as an additional coordinate rather than a feature.

To quantify the performance of our approach, we apply the $F_1$-Score as the harmonic mean of precision and recall. As depicted in Table 2, we calculate those statistics for each class individually and then determine the average scores for precision, recall and $F_1$ to provide a general performance measure for each approach. First, we use a vanilla version of PointCNN (second row) for the semantic segmentation task and compare it to PointNet ++ (first row) as defined in [18]. This comparison clearly demonstrates the superior performance of PointCNN on detecting moving objects. In addition, Table 2, shows a discrepancy in memory and computing power requirements between these two algorithms. While PointCNN uses parameter-expensive X-Convolutions, PointNet++ applies shared MLPs as a feature extractor which leads to an extensive number of FLOPs needed for a forward pass of the network.

Furthermore, Table 2 reveals the value of the pre-processing network as well as MSG to adapt PointCNN to radar data. First, applying pre-processing to encode features of our initial radar PC increases the overall performance of PointCNN while adding a slight cost in terms of memory and computation. Replacing k-nearest neighbour with MSG, however, improves the capability of the network to cluster detections of large size objects like vehicles. When applied without PP, it slightly decreases the segmentation performance on pedestrians. Finally, when we combine pre-processing with MSG in PointCNN, the resulting algorithm outperforms any of those modification implemented separately as well as the vanilla version of PointCNN. Further, when compared to PointNet++ as proposed in [18], our approach of combining PP with MSG achieves an average improvement of ~4% in $F_1$-Score.

PointNet++ is widely used as a basic component to extract descriptive patterns from unordered data formats like PCs. Even PCNN [7], the so far best performing network for the semantic segmentation of radar PCs, uses PointNet++ as a central component for feature extraction. When compared in Table 2 our novel version of PointCNN outperforms RadarPCNN by a margin of ~1% in $F_1$-Score for the moving vehicle class and a slight margin in average $F_1$-Score. Therefore, our approach not only offers an alternative to PointNet++ but also reveals a superior network architecture for the semantic segmentation of radar PCs.

**Table 2.** Semantic segmentation results (%).

| | Moving Vehicle | | | Moving Pedestrian | | | Average | | | | |
|---|---|---|---|---|---|---|---|---|---|---|---|
| Method | Prec. | Recall | $F_1$ | Prec. | Recall | $F_1$ | Prec. | Recall | $F_1$ | Param. | FLOPs |
| **PointNet++** [18] | 72.77 | 82.89 | 77.50 | 46.10 | 75.77 | 57.32 | 59.44 | 79.33 | 67.41 | 418.6K | 1.55G |
| **PointCNN** | 76.48 | 87.59 | 81.66 | 45.57 | 75.71 | 56.89 | 61.3 | 81.65 | 69.28 | 653.2K | 0.84G |
| **PointCNN (PP)** | 79.30 | 85.07 | 82.08 | 46.58 | 76.55 | 57.92 | 62.94 | 80.81 | 70.00 | 663.8K | 0.86G |
| **PointCNN (MSG)** | 80.45 | 84.85 | 82.59 | 43.71 | 75.11 | 55.26 | 62.08 | 79.98 | 68.93 | 653.2K | **0.83G** |
| **RadarPCNN** [7] | 75.78 | **91.44** | 82.88 | **48.64** | 75.82 | **59.27** | 62.21 | **83.63** | 71.07 | **175.3K** | 1.02G |
| **PointCNN (PP+MSG)** | **80.81** | 87.84 | **84.18** | 47.88 | 75.35 | 58.55 | **64.35** | 81.60 | **71.37** | 663.8K | 0.85G |

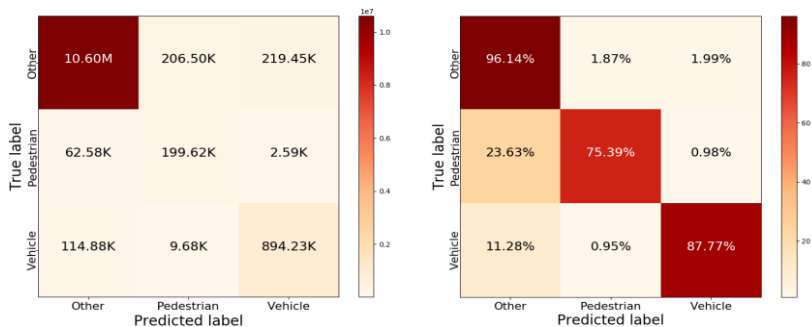

**Figure 3.** Confusion matrices of predicted classes using our optimized version of PointCNN. On the left side we show the absolute amount of predictions, on the right side the point counts in relation to all labeled detections of a class are presented.

Confusion matrices from Figure 3 provide an extensive insight into the performance of our approach. In the matrix on the left, a large imbalance can be observed between detections for the pedestrian / vehicle and the static background class, with a large excess of occurrence for the latter. Even though this effect should be compensated for by Focal Loss mechanisms that we built in, a tendency for the network to predict detection as background cannot be completely ruled out. This can further be discovered from the matrix on the right side of Figure 3. Here, we obtain a major confusion between detections from vehicles and pedestrians that are falsely predicted as background. We belief that besides the bias towards the background class that we already mentioned, this behaviour can be traced back to the discrepancy in our data between radial doppler speed and the absolute speed of detected objects which makes it hard for the network to rely on doppler signatures. On the other hand, our approach shows robust characteristics in distinguishing between pedestrians and vehicles with a minor false positive rate of ~1%, respectively.

Figure 4 emphasizes the value of radar-specific, intrinsic properties $v_r$ and $\sigma$ of each detection in the PCs. The figure shows that removing $\sigma$ as an input property of the network only marginally decreases the network performance represented by the F1-Score (blue curves) for both vehicle and pedestrian reflections. On the other hand, removing $v_r$ from the input feature vector causes a huge drop in performance. This shows the significant importance of Doppler velocities in classifying detections.

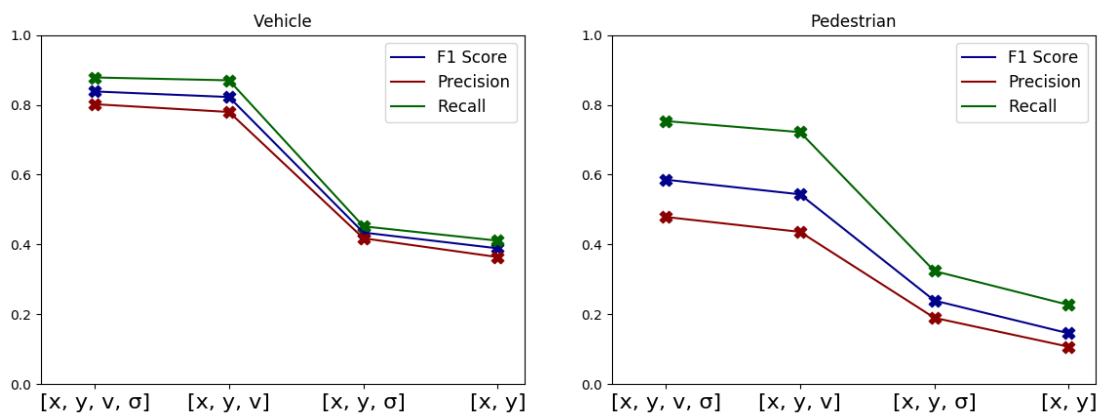

**Figure 4.** Ablation studies on radar PC properties that the network consumes for semantic segmentation. F1-Score as well as Precision and Recall are shown for feature vectors with subsequent removing $v_r$ and $\sigma$ until semantic segmentation is carried out exclusively based on the coordinates x and y.

## 5. Conclusion and future work
In this work we show how PointCNN can effectively be applied for pattern recognition within radar PCs, outperforming previous approaches that are built upon PointNet++ on the task of semantic segmentation. We exploit the strengths of PointCNN to learn how to pre-process a cluster of radar detections with the help of neural networks so that detections within the cluster can be convolved to extract spatially local correlations. We then implement some tweaks on PointCNN to adjust the algorithm to the characteristics of radar data like low density, heterogenous distributed detections and intrinsic properties of each radar reflection. Therefore, we first apply the idea of a pre-processing network to cast inputs x, y, doppler velocity and radar cross section of our initial radar PC to a representation that makes it easier for the network to extract patterns. In addition, we improve clustering of detections within PointCNN by implementing multi-scale grouping to ensure clusters of neighbouring radar detections. When applied on the semantic segmentation task, our network reaches superior performance compared to previous approaches.

The ideas presented in this work can be used to replace PointNet++ as a feature extractor in RadarPCNN. In this way, the attention mechanism of RadarPCNN could be combined with the superior performance of PointCNN on convolving PCs.